\pdfoutput=1

\documentclass[11pt]{article}

\usepackage[]{emnlp2021}

\usepackage{times}
\usepackage{latexsym}
\usepackage{url}
\usepackage{adjustbox}
\usepackage[T1]{fontenc}
\usepackage{subcaption}
\usepackage{algorithm}
\usepackage{algorithmicx}
\usepackage{xcolor}
\usepackage{graphicx}
\usepackage{amsmath}
\usepackage{dsfont}
\usepackage{pgfplots}
\usepackage{booktabs}
\usepackage[utf8]{inputenc}
\usepackage{microtype}

\usepackage{xspace,mfirstuc,tabulary}

\newcommand{\com}[1]{}

\title{Improved Goal Oriented Dialogue via Utterance Generation and Look Ahead}

\author{Eyal Ben-David\Thanks{Both authors equally contributed to this work.} \Thanks{Work done during an internship at IBM Research.} \textsuperscript{1}  , Boaz Carmeli* \textsuperscript{2}, Ateret Anaby-Tavor\textsuperscript{2}\\
  \textsuperscript{1}Faculty of Industrial Engineering and Management, Technion, IIT \\
  \textsuperscript{2}IBM Research \\
  \texttt{eyalbd12@campus.technion.ac.il} \\
  \texttt{\{boazc,atereta\}@il.ibm.com} \\}

\begin{document}

\maketitle

\begin{abstract}
Goal oriented dialogue systems have become a prominent customer-care interaction channel for most businesses.
However, not all interactions are smooth, and customer intent misunderstanding is a major cause of dialogue failure.
We show that intent prediction can be improved by training a deep text-to-text neural model to generate successive user utterances from unlabeled dialogue data.
For that, we define a multi-task training regime that utilizes successive user-utterance generation to improve the intent prediction.
Our approach achieves the reported improvement due to two complementary factors: First, it uses a large amount of unlabeled dialogue data for an auxiliary generation task. Second, it uses the generated user utterance as an additional signal for the intent prediction model.
Lastly, we present a novel look-ahead approach that uses user utterance generation to improve intent prediction in inference time.
Specifically, we generate counterfactual successive user utterances for conversations with ambiguous predicted intents, and disambiguate the prediction by reassessing the concatenated sequence of available and generated utterances.

\end{abstract}

\section{Introduction}
\label{sec:intro}

Dialogue systems have emerged as a prevalent method for humans to interact with their digital surroundings \citep{chen2017survey:41, ralston2019voice:46, bocklisch2017rasa:27}.
Consequently, goal-oriented dialogue systems have become a prominent customer-service interaction-channel for most businesses \citep{yan2017building:44}.
However, there is still a long way to go until we reach a pervasively-integrated, fully-automated, and reliable dialogue system to serve as the preferred customer-care go-to channel \citep{gao2018neural:16}.

\begin{figure}
\includegraphics[scale=0.39]{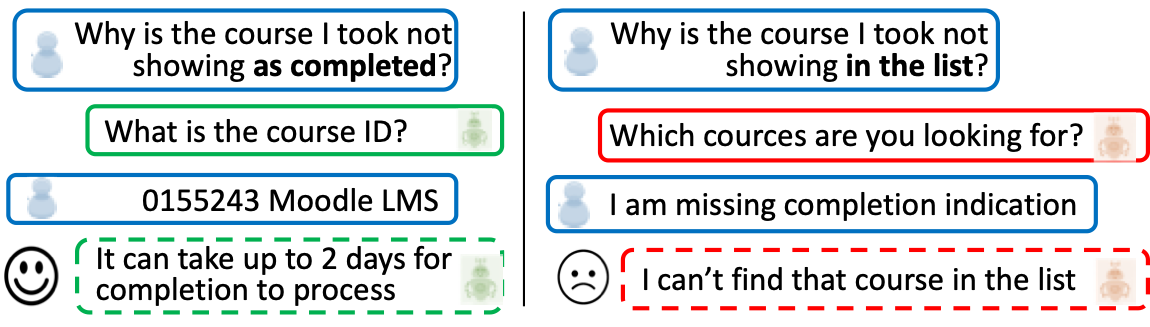}
\centering
\caption{Using dialogue context for improved intent prediction. Left box: Successfully completed dialogue. Right box: Bot misunderstood user intent thus provided an incorrect response. Error is attributed to the incorrect attention that the dialogue system gives to the misleading adjective 'in the list' instead of the indicative verb 'took'.}
\centering
\label{fig:good-bad-dialogue}
\end{figure}

A key component of these systems are information-completion processes that are mutually carried out by two actors: the user and the bot \citep{luo2014win:45}. 
The user knows her goal but is usually unaware of the various options and the optimal way to achieve it.
The bot, on the other hand, has extensive knowledge of the possible options and the various ways to achieve them, but is unaware of the user's exact goal.
Via a dialogue, the two actors cooperate to achieve the user's goal in the most effective way.

Typically, users express their intent at the beginning of the conversation, and the bot reacts with its most suitable response.
Bot responses are dependent on user intents.
Ultimately, an accurate bot response depends on its ability to correctly understand the intent.
Bot misunderstanding of the user intent will most probably lead to an unsatisfactory response, which in turn will cause the user to react accordingly (see Figure \ref{fig:good-bad-dialogue}).
For example, the user might ask for a human agent, express negative sentiment, or rephrase her request. 
Therefore, the likelihood of a dialogue system to retrospectively identifying errors in its initial predicted user intent increases as the conversation progresses.

Given this observation, we hypothesise that \textbf{successive user utterances provide an invaluable, yet overlooked, indication} for the language understanding level expressed by the previous bot response.
Thus, training a model to predict successive user utterances can help predict the user's intent (\S~\ref{sec:user-side-modulation}, \ref{sec:utterance-generation}).

To validate our hypothesis, we experiment with training a text-to-text transformer model using multi-task regime that simultaneously i) generates predicted user utterances from a vast amount of unsupervised dialogue data; and ii) predicts the intent from dialogue context, enriched with successive, generated, user utterances.
While earlier deep neural network models require a dedicated classification layer and parameters \citep{devlin2018bert}, a text-to-text model, such as T5 \citep{DBLP:journals/corr/abs-1910-10683}, solves the classification task in a generative manner.
This approach allows the model to share all its parameters across the prediction and generation tasks and thus, potentially achieve better results (\S~\ref{sec:transformer-lms}).

Having a model that is trained to generate user utterances suggests an innovative way to further improve the intent prediction by generating successive user utterances and using them as a look-ahead signal during inference (\S~\ref{sec:utterance-generation-task}). Furthermore, such a model can also assist in solving the specific situation of conflicting intents.
For that, one may generate counterfactual successive user utterances and use them as a look-ahead signal to disambiguate the conflicting intents (\S~\ref{sec:ablation}).


Our contribution is thus, two-fold. We show that i) training a text-to-text model using a multi-task regime, in which the secondary task is to predict the succeeding user utterance, further improves the main task of intent prediction, and ii) generating a successive user utterance and using it as a look-ahead signal improves prediction in inference time.



\section{Related Work}
\label{sec:related-work}

\subsection{Contextual Intent Prediction}
Intent prediction is usually formulated as a sentence classification task. Given an utterance (e.g., “what are the details of this flight?”), a system needs to predict its intent (e.g., “flight details”). Modern approaches use neural networks to model intent prediction, using many different classification architectures such as RNNs \citep{DBLP:conf/interspeech/RavuriS15}, CNNs \citep{ DBLP:conf/nips/ZhangZL15}, attention-based CNN \citep{DBLP:conf/interspeech/ZhaoW16}, and transformer-based architectures \citep{devlin2018bert}.  
In many cases, intent prediction is modeled jointly with slot filling \citep{DBLP:conf/asru/XuS13, DBLP:conf/interspeech/ChenHTGD16, DBLP:conf/interspeech/LiuL16, DBLP:conf/naacl/GooGHHCHC18, DBLP:journals/corr/abs-1902-10909}. 

Another way to approach this prediction task is to use more contextual knowledge, such as more dialogue turns \citep{feng2020regularizing:51}. This approach was made possible thanks to the creation of public datasets for task-oriented dialogue, which contain utterance level annotations for full conversations \citep{budzianowski2018multiwoz:2, DBLP:conf/aaai/RastogiZSGK20}. While most studies in this direction seek to build a dialogue state tracking model \citep{DBLP:conf/interspeech/LiuL17, DBLP:journals/corr/abs-1812-00899, cheng2020conversational:50}, several works model it together with intent prediction \citep{DBLP:journals/corr/abs-1912-09297}.  

In this paper, we use the dialogue context for intent prediction. Unlike previous work, we face a setup in which there are few conversations with intent-annotation, as we consider this more realistic for a system's development phase.
Furthermore, we use only dialogue utterances for prediction, without any other information (e.g., internal dialogue state tracking variables).      

\subsection{Modeling User-side Utterances}
\label{sec:user-side-modulation}
Modeling the user side in dialogue systems has been researched in several tasks and applications. One approach is to develop a user simulator, which is essential e.g., for training dialogue models based on reinforcement learning (RL). Naturally, the performance of the simulator directly impacts the RL policy. Within this line of work, two methods have proven particularly useful: 1) Agenda-based simulators, using rule-based software to generate a simulation \citep{DBLP:journals/ker/SchatzmannWSY06, DBLP:conf/naacl/SchatzmannTWYY07, DBLP:journals/corr/LiLDLGC16}, and 2) data-driven simulators that learn user responses directly from a corpus \citep{DBLP:conf/interspeech/AsriHS16, DBLP:conf/emnlp/HeCBL18, DBLP:conf/sigdial/KreyssigCBG18}. 

The modeling of users' written and spoken utterances is another direction that has been extensively studied within the standard dialogue modeling framework  \citep{DBLP:books/lib/KobsaC89, DBLP:journals/ais/Kobsa90, DBLP:conf/aiide/LinW11, DBLP:conf/acl/LiGBSGD16}.
Existing methods strive to improve users' language consistency in conversations and generate meaningful responses in an open-domain dialog. 

In this work, we use user-side modeling to improve the performance of goal-oriented dialogue systems.
We rely on the hypothesis that the user conveys her intent to the bot and then implicitly reflects, in successive utterances, whether the bot correctly understood it.
Our model, thus, incorporates the generation of successive user-utterances, aiming to learn from this weak-supervision signal.

\subsection{Transformer-based Language Modeling}
\label{sec:transformer-lms}
Previous research works consider two main approaches when training a language model (LM) that is based on the Transformer architecture \citep{DBLP:conf/nips/VaswaniSPUJGKP17}.
The first treats the model as a contextualized word embedding (CWE) \citep{devlin2018bert, yang2019xlnet, liu2019roberta, DBLP:journals/corr/abs-1910-01108, DBLP:conf/acl/MartinMSDRCSS20}. As such, it employs the model as an encoder. When fine-tuning the model, additional architecture is placed on top of the model, adjusting model's CWE representation to a specific downstream discriminative task.

The second approach handles the task in a sequence-to-sequence manner.
As such, it uses an encoder-decoder architecture (or just a decoder \citep{radford2019language}), such that when given a context, it generates the missing tokens.
Because of their generative nature, these models are commonly integrated into text generation downstream tasks \citep{radford2019language, DBLP:conf/aaai/Anaby-TavorCGKK20, DBLP:journals/tacl/RotheNS20}.
Recently, \citet{DBLP:journals/corr/abs-1910-10683} presented T5, a text-to-text transfer transformer, and showed that encoder-decoder LMs are efficient in both discriminative and generative tasks.
\textbf{T5} demonstrated its superiority in many tasks, eliminating the need to add task-related architecture.


In this work we follow the second approach and take advantage of the T5 unified framework, which converts all text-based language problems into a text-in text-out format.
We apply T5 to the task of intent prediction under the low-supervision scenarios we encounter by
enriching our supervised task with semi-supervised signals from unlabeled data.

\subsection{Utterance Generation for Look Ahead}
\label{sec:look-ahead-prev}
Although there is line of work on learning from user feedback \citep{hancock2019learning:6, shin2019happybot:48}, and specifically from generating user-side utterances \citep{weston2016dialog:13}, ours is the first work we know of that uses generated user utterance as a look-ahead signal to improve dialogue understanding during inference.

\section{Model and Methods}

\begin{figure}[t!]
\includegraphics[scale=0.45]{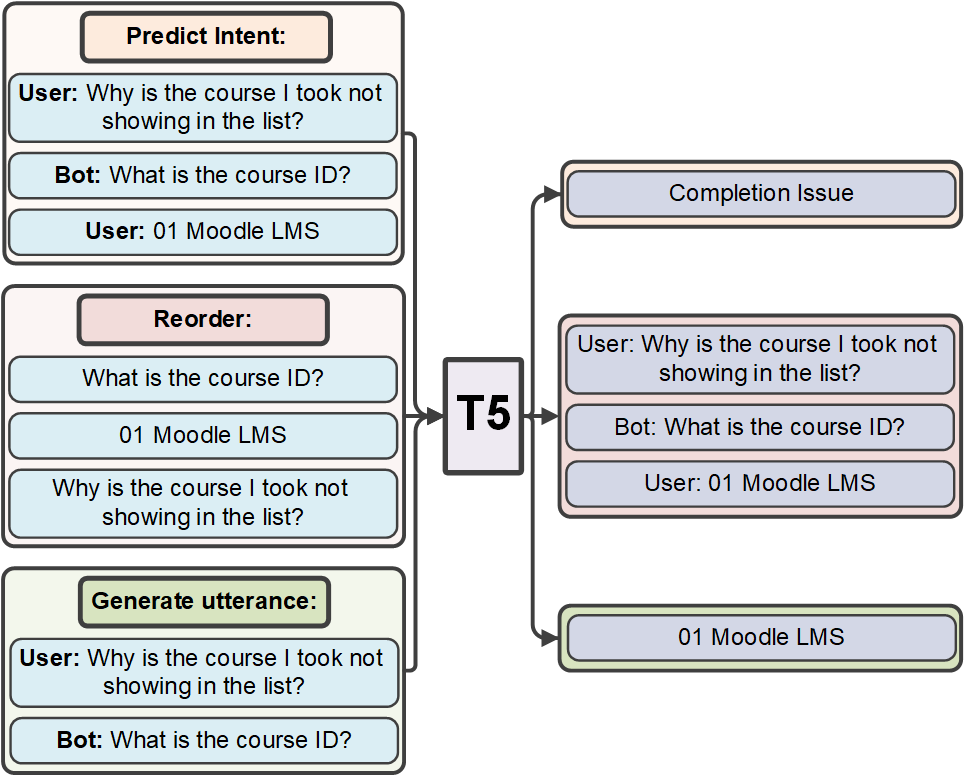}
\centering
\caption{T5 model trained on three tasks: 1) predict intent, 2) reorder utterances and 3) generate third utterance (3UG).}
\centering
\label{fig:model}
\end{figure}

Dialogue representation relies on a complex data structure, composed of multiple utterances, generated alternately by two actors: user and bot.
The number of utterances in a dialogue varies dramatically and so does the length of each utterance.
To cope with this complexity we based our experiments on a flexible text-to-text model (T5) that supports multi-task training by design via encoder-decoder architecture and language-modeling training. 
In T5, all tasks are mapped into text-in text-out format.
Specifically, the model resolves classification tasks by generating the predicted class' label.
The model is thus able to use the same loss function and parameter weights across different tasks, data sets, and training regimes.

Next, we explain our text-in text-out experimental method in more details.

\subsection{Contextual Intent Prediction}
\label{sec:intent-prediction-task}
We start by training two T5-based variants:
The first, referred to as semantic \textit{utterance} classification (\emph{SUC}) \citep{tur2012towards:43} is trained on just the first user-utterance from each dialogue. This classifier serves as our baseline.
The second, referred to as semantic \textit{dialogue} classification (\emph{SDC}), is trained on, at least, three successive dialogue utterances. 

For low data regimes, we further enriched the supervised dialogue data with a weak supervision signal.
To that end, we trained RoBERTa \citep{liu2019roberta} and BERT \citep{devlin2018bert}, two state-of-the-art classifiers, on the available supervised dataset and used them to predict the intent for unlabeled conversation.
We added samples for which the two classifiers agreed on the predicted intent to the supervised train set.

To measure the model's ability to correctly predict the intent, we constructed a full-dialogue test set.
Each dialogue in this set consisted of at least three successive utterances that pertain to the same user intent.
We measured the model's accuracy by classifying one, two, and three initial utterances from each dialogue in the dataset.  

\subsection{Utterance Generation}
\label{sec:utterance-generation}
To evaluate the hypothesis that utterance generation improves intent prediction, we defined a generative task to predict (i.e., by generating) the third user utterance from a given first user utterance and successive bot response. 
We refer to this task as \emph{3UG}.
To train the T5 model on this task, we used a vast amount of unlabeled data that was continuously logged by the dialogue system during interaction with users.
Alternatively, dialogue data from user to human-agent interactions can be used in situations where data from dialogue systems does not exist.

\begin{figure}
\includegraphics[scale=0.37]{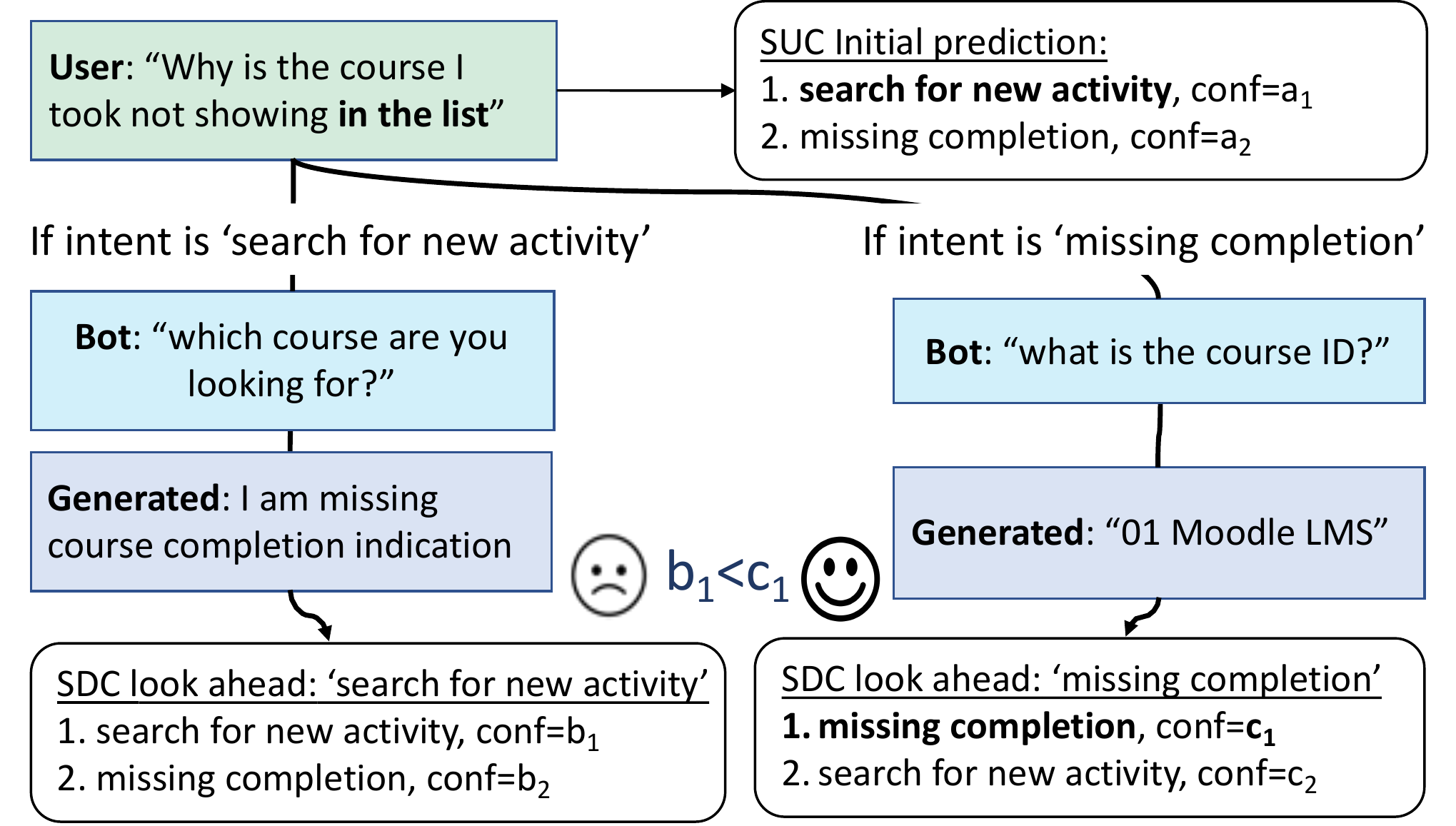}
\centering
\caption{Using third utterance generation to improve intent prediction via look ahead. $a_1 > a_2$ are \emph{SUC} confidence scores when using just the first utterance. $b_1 > b_2$ are \emph{SDC} confidence scores when using three utterances and assuming that \textit{search for new activity} is the intent. $c_1 > c_2$ are \emph{SDC} confidence scores when using three utterances and assuming \textit{missing completion} is the intent. In this example, the highest score is $c_1$, thus the predicted intent is \textit{missing completion}.}
\centering
\label{fig:counterfactual-scenario}
\end{figure}

We test results by comparing the \emph{SDC} performance after training on the \emph{3UG} task to the \emph{SUC} intent prediction performance.
Importantly, one may notice that the \emph{SDC} model takes advantage of the additional supervision signal available within the successive user utterance during training, although this utterance is not available during inference.
Thus, for comparing classifier performance on real-time conversations, we experiment with complementing the first available utterance with two additional utterances as follows.
First, we used the dialogue system to get the bot response i.e., the second utterance.
Second, we used the T5 model already-trained on the \emph{3UG} task to generate a successive user utterance, conditioned upon the first user utterance and the bot response (see Figure \ref{fig:counterfactual-scenario}).

We report results for these two complementary experiments in section \ref{sec:results}. 

\subsection{Disambiguate Conflicting Intents}
\label{sec:utterance-generation-task}
Conflicting-intents refers to situations where the classifier prediction doesn't clearly distinguish between small set of competing intents. 
To solve this specific problem we experimented with a novel counterfactual approach that uses the \textit{3UG} model to generate a counterfactual user utterance for each of the conflicting intents as described in \S~\ref{sec:utterance-generation}.
We then evaluated the performance by predicting the intent based on the ensemble of predictions for all look-ahead conversations, where each conversation now has three successive utterances.




\subsection{Multi-task Training Regimes}
\label{sec:training-regimes}
We wanted to evaluate the level of intrinsic synergy between intent classification and utterance generation tasks.
To accomplish this, we compared the intent prediction accuracy achieved by T5 trained on the supervised full-dialogue dataset alone, (thus performing only a classification task), to the results when the model was asked to first perform the successive utterance generation task (trained on the unsupervised dataset) and then predict the intent (trained on the supervised dataset).
We also extracted an \textit{utterance reordering} task out of unsupervised data, in which we shuffled dialogue utterances and train the model to correctly reorder them. We elaborate more on dialogue-related auxiliary tasks in the supplemental material. 

\section{Experiments}
\label{sec:experiments}
We conducted most of the experiments by fine-tuning a pre-trained T5-base model, which shares its parameters across tasks.
Specifically, we fine-tuned the T5 model repeatedly using unsupervised, semi-supervised, and supervised datasets in various orders across three different datasets.
\subsection{Datasets}
\label{sec:Dataset}
We used the following datasets (see Table \ref{tab:datasets}):\\
\textbf{\textit{MultiWOZ}} - a large-scale multi-domain Wizard-Of-Oz dataset for task-oriented dialogue modelling, containing over 8400 dialogues, and spanning seven domains \citep{budzianowski2018multiwoz:2}.
This dataset contains human to human conversations, and thus does not obey any well-defined bot-side dialogue rules or learning algorithm.\\
\textbf{\textit{SGD}} - a schema-guided dialogue dataset, containing over 16K conversations, spanning 86 intents, and covering 16 different domains \citep{rastogi2019towards:9}.
This dataset was created by asking crowd-sourced workers to rephrase dialogue utterances that were created by a rule-based simulator.
Thus, this dataset obeys predefined dialogue rules but still reflects the language diversity and variations one expects to find in goal-oriented dialogues.\\
\textbf{\textit{EDU}} - an unpublished dataset extracted from an operational goal-oriented dialogue system.
It contains over 4K dialogues across 115 intents pertained to the learning-and-education domain; out of these 4K dialogues, only 350 have intent labels.
This dataset was extracted from a real customer-care dialogue system that uses an SVM-based intent classifier and a rule-based dialogue-flow software.
Intent identification in this dataset depends solely on prediction from the SVM-based single-utterance classifier; the dialogue system uses additional rule-based logic to return a concrete bot response.

From the original datasets, we extracted conversations with at least three successive utterances pertaining to the same intent.
We further split dialogues that expressed multiple intents into different parts, keeping those that have at least three successive utterances pertaining to a single intent as separate dialogues.
This left us with 3994, 34056, and 2063 conversations pertaining to 11, 86, and 115 intents for \textit{MultiWOZ}, \textit{SGD}, and \textit{EDU}, respectively.
From these we used 1012, 1000 and 100 conversations for supervised training and 2,538, 23,128, and 1,200 conversations for unsupervised training, for the three datasets, respectively.
For \textit{MultiWOZ} and \textit{SGD}, we mimicked the situation of unsupervised data by ignoring the available label.
For \textit{EDU}, we had only 350 labeled conversations, which we split into 100, 50, and 200 for the train, validation and test sets, respectively.  
To increase the very low amount of supervised \textit{EDU} data, we further applied the weak-labeling algorithm described in \S~\ref{sec:intent-prediction-task}, which added an additional 513 semi-supervised conversations.

\begin{table}[t]
\small
\centering
\begin{tabular}{| l | c | c | c |}
\hline
   & \textbf{\textit{MultiWOZ}} & \textbf{\textit{SGD}} & \textbf{\textit{EDU}} \\
    \hline
    \#Intents  & $11$ & $86$ & $115$ \\
    \hline
    Unsupervised  & $2,538$ & $23,128$ & $1,200$ \\
    Supervised  & $1,012$ & $1,000$ & $100$ \\
    Weakly labeled  &  $0$ & $0$ &  $513$ \\
    \hline
    Dev  & $211$ & $4933$ & $50$ \\
    Test  & $233$ & $4995$ & $200$ \\
    \hline
     
\end{tabular}
\caption{Intents and data-splits for the three datasets}
\label{tab:datasets}
\end{table}

\subsection{Models and Baselines}
\label{sec:models}
We used \emph{SDC} as our main model for researching the performance of contextual intent prediction.
This involved training \emph{SDC} by fine-tuning a pre-trained T5 model on the supervised data. 

We then used \emph{SUC} as a baseline model for the contextual intent prediction experiment.
To train \emph{SUC}, we fine-tuned a pre-trained T5 model with supervised data containing just the first utterance from each dialogue and the associated intent label.


To better understand the effect of the multi-task training, we further examined three \emph{SDC} model variants that differ in their fine-tuning regimes:\\
\textbf{\emph{ALL}}, which we fine-tuned on all available datasets: the unsupervised, supervised, and semi-supervised, as described in \S~\ref{sec:training-regimes}.\\
\textbf{\emph{PART-SDC}}, which we fine-tuned in two steps: first on all except for the supervised intent prediction dataset, and then on the supervised dataset.\\
\textbf{\emph{ALL-SDC}}, which we also fine-tuned in two steps: first on all available datasets and then, again, on the supervised intent prediction dataset. 

\subsection{Experimental Setup}
\label{sec:experimental=setting}
Our experiments assume at least three consecutive dialogue utterances of user-bot-user relating to the same intent.
We tested model's performance in predicting the user intent as the dialogue progressed by letting it classify one, two, and three consecutive utterances from each dialogue available in the test set.
We refer to these as \emph{1-u}, \emph{2-u} and \emph{3-u} test scenarios, respectively, as can be seen in Table \ref{tab:main-table}.
To evaluate performance, we measured intent accuracy, namely $t/d$ where $t$ is the number of true predictions and $d$ is the number of samples, i.e., predictions, in the test data. In the \textit{EDU} experiments, we report a model's improvement in accuracy with respect to \emph{SUC}, to avoid revealing sensitive information regarding an inner-system's performance.

To evaluate the hypothesis that a third generated utterance serves as a good look-ahead signal during run-time, we sampled such utterances from the best-performing T5 model and used them instead of the actual utterance available in the test set.
This experiment is referred to as \emph{3-gen} in Table \ref{tab:comp-generation-lookahead}.
To better understand the power of the generative model to produce qualitative utterances, we experimented with the 'five times third-user-utterance' method, referred as  \emph{3-5xg}.
This method generates five third-user-utterance alternatives for each two successive user-bot utterances.
It then predicts the intent when each one of them is used as a third utterance.
Last, it uses majority voting to select from the predicted intents.
We complement these two generation methods with a baseline method, referred as \emph{3-rnd}, which generated a random utterance and used it to predict the intent. 


\section{Results}
\label{sec:results}
We report the key results of our experiments in Table \ref{tab:main-table}.
As shown, the performance of intent prediction increases significantly with dialogue progression along all datasets, even if the model was trained on just the supervised data (\emph{SDC}).
Specifically, \textit{MultiWOZ}: $93.1\% \rightarrow 96.6\%$; \textit{SGD}: $68.5\% \rightarrow 73.6\%$; \textit{EDU}: $+1.8\% \rightarrow +6.3\%$.
Moreover, training the model on all available data and all tasks (\emph{ALL-SDC}) improved the results across all datasets and all test scenarios: \emph{1-u}, \emph{2-u}, and \emph{3-u}.
These results clearly prove our hypothesis that generating a third utterance as a secondary task in a multi-task training regime improves intent prediction performance.

Interestingly, we observed that both \emph{SDC} and \emph{ALL-SDC} models outperform the baseline \emph{SUC} model even when tested on the \emph{1-u} scenario, namely when tested to predict the intent from only the first utterance.
This phenomenon occurs across all datasets (\textit{MultiWOZ}: $90.5\% \rightarrow 93.1\% \rightarrow 94.4\%$; \textit{SGD}: $68.2\% \rightarrow 68.5\% \rightarrow 71.1\%$; \textit{EDU}: $0.0\% \rightarrow +1.8\% \rightarrow +5.5\%$).
This important finding clearly indicates the benefit of training text-to-text models over full conversations, even if the model is used to predict intents from just the first utterance during inference. 


\begin{table}[t]
\small
\centering
\begin{tabular}{c | c | c | c | c }
\toprule
    \textbf{\textit{MultiWOZ}} & \emph{1-u} & \emph{2-u} & \emph{3-u} & \emph{3-5xg}\\
    \hline
     \emph{SUC} & 90.5 & - & - & -\\ 
     \hline
     \emph{SDC} & 93.1 & 95.5 & 96.6 & 96.6 \\
     \emph{ALL-SDC} & 94.4 & 97.2 & \textbf{98.7} & 98.3\\
     \toprule
     \textbf{\textit{SGD}} & \emph{1-u} & \emph{2-u} & \emph{3-u} & \emph{3-5xg}\\
     \hline
     \emph{SUC} & 68.2 & - & - & -\\ 
        \hline
     \emph{SDC} & 68.5 & 73.2 & 73.6 & 74.0\\
     \emph{ALL-SDC} & 71.1 & 74.2 & 74.6 & \textbf{75.8}\\
     \toprule
    \textbf{\textit{EDU}} & \emph{1-u} & \emph{2-u} & \emph{3-u} & \emph{3-5xg}\\
     \hline
     \emph{SUC} & $0.0$ & - & - & -\\ 

        \hline
     \emph{SDC} & $+1.8$ & $+1.8$ & $+6.3$ & $+5.5$\\

     \emph{ALL-SDC} & $+5.5$ & $+8.2$ & $\textbf{+9.1}$ & $+8.2$\\



\end{tabular}
\caption{Results for three different models (\emph{SUC}, \emph{SDC}, \emph{ALL-SDC}) on four different test scenarios (\emph{1-u}, \emph{2-u}, \emph{3-u}, \emph{3-5xg}) across the three datasets.}
\label{tab:main-table}
\end{table}

\paragraph{Improved prediction via look-ahead}
\label{par:utteranc-generation}
The rightmost column (\emph{3-5xg}) in Table \ref{tab:main-table} reports results of the look-ahead experiment.
Here, we generated five user utterances and used majority voting to choose among them.
As can be observed, the generation of a third utterance improves intent prediction compared to using just the two utterances available during inference.
Notably, these results hold for all datasets and all models except \textit{EDU}:\emph{ALL-SDC}.
Interestingly, intent classification performance with the generated utterance (\emph{3-5xg} column) is comparable to the one using real utterances (\emph{3-u} column).
More specifically, \textit{MultiWOZ}: $98.7\%$ vs $98.3\%$; \textit{SGD}: $74.6\%$ vs $75.8\%$; \textit{EDU}: $+9.1\%$ vs $+8.2\%$.

\section{Ablation Analysis}
\label{sec:ablation}
We conducted several ablation experiments to shed more light on our key results. 
First, we investigated the impact of various training regimes on model performance.
Next, we examined the effect of conversation reordering as an auxiliary task in our multi-task training regime.
Then, we assess the quality of utterances sampled from our generative models.
Finally, we demonstrate the ability of our model to generate counterfactual utterances which may resolve situations of conflicting intents. 
To ensure compatibility with previous results, we kept all experiment details similar to what described in \S~\ref{sec:experimental=setting}.   

\begin{table}
\small
\centering
\begin{tabular}{c | c | c | c}
\toprule
     \textbf{\textit{MultiWOZ}} & \emph{1-u} & \emph{2-u} & \emph{3-u}\\
     \hline
     \emph{SDC} & 93.1 & 95.5 & 96.6\\
     \emph{PART-SDC} & 94.0 & 95.7 & 97.0\\
     \emph{ALL} & 94.8 & 97.0 & 98.2\\
     \emph{ALL-SDC} & 94.4 & 97.2 & 98.7\\
     \toprule
     \textbf{\textit{SGD}} & \emph{1-u} & \emph{2-u} & \emph{3-u}\\
     \hline
     \emph{SDC} & 68.5 & 73.2 & 73.6\\
     \emph{PART-SDC} & 68.7 & 73.3 & 74.0\\
     \emph{ALL} & 71.6 & 76.1 & 76.6\\
     \emph{ALL-SDC} & 71.1 & 74.2 & 74.6\\
     \toprule
     \textbf{\textit{EDU}} & \emph{1-u} & \emph{2-u} & \emph{3-u}\\
     \hline
     \emph{SDC} & $+1.8$ & $+1.8$ & $+6.3$\\
     \emph{PART-SDC} & $+0.9$ & $+2.7$ & $+5.4$\\
     \emph{ALL} & $-1.8$ & $+2.7$ & $+6.3$\\
     \emph{ALL-SDC} & $+5.4$ & $+8.2$ & $+9.1$\\
\end{tabular}
\caption{Results from four different training regimes, tested for accuracy on one, two and three dialogue utterances across the three datasets.}
\label{tab:training-regimes-table}
\end{table}

\paragraph{Training regimes effect}
Table \ref{tab:training-regimes-table} presents results from the three different training regimes \emph{PART-SDC, ALL, ALL-SDC}, compared to the \emph{SDC} baseline-model trained solely on the supervised data.
Adding unsupervised and semi-supervised data was shown to improve the \emph{SDC} model performance across all training regimes.
Further analysis shows that fine-tuning the model in two steps, first on all but the supervised data and then on the supervised data (\emph{PART-SDC}), is inferior to fine-tuning the model on all available data in a single step (\emph{ALL}).
Lastly, we observe that as in \emph{ALL-SDC}, the two-step training regime further improves results over the \emph{ALL} model when tested on three utterances, in two out of the three datasets (\textit{MultiWOZ}: $98.2\% \rightarrow 98.7\%$; \textit{EDU}: $+6.3\% \rightarrow +9.1\%$)

\begin{figure}[t]
\includegraphics[scale=0.13]{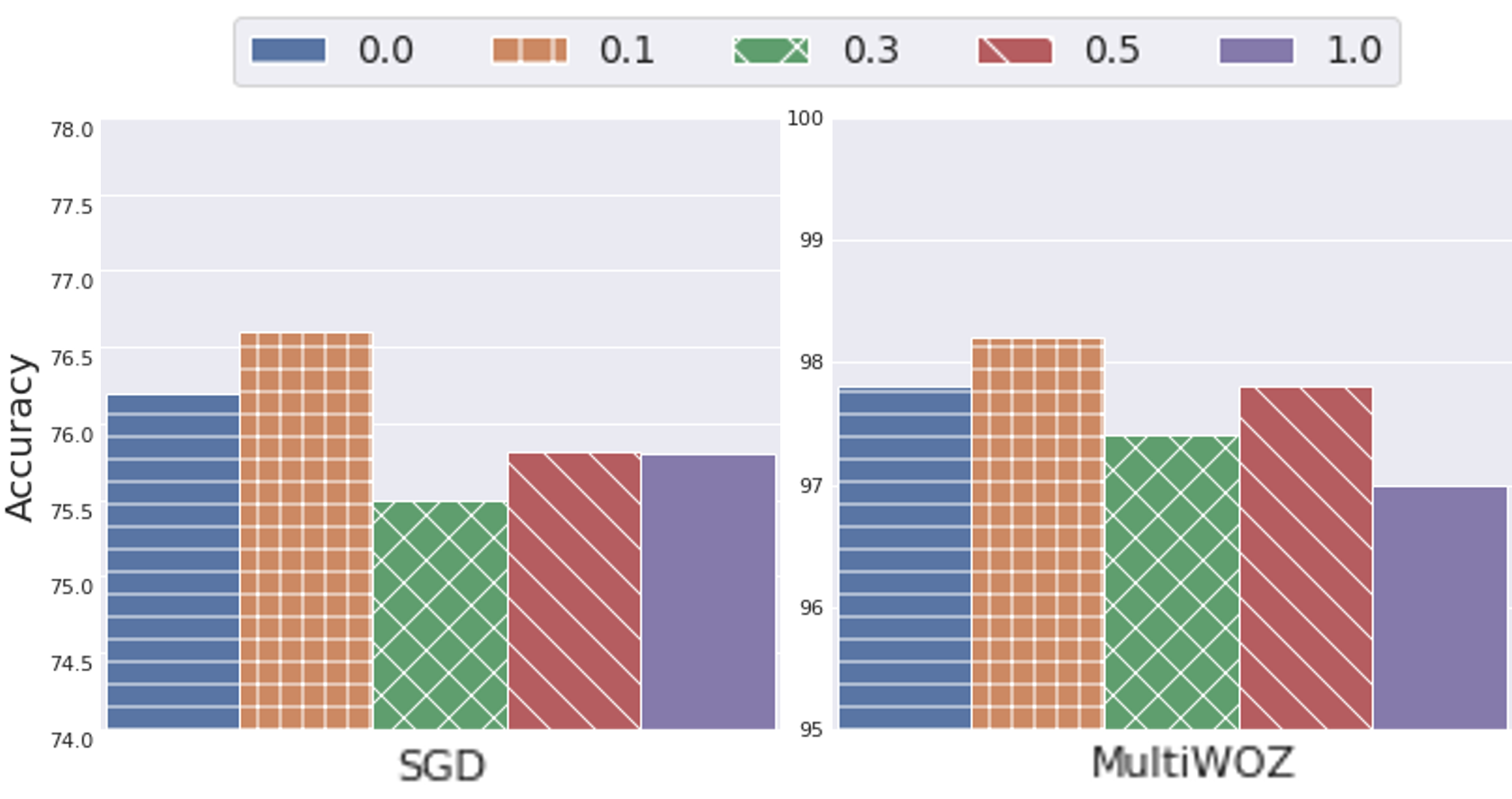}
\centering
\caption{The data proportion for reordering conversations according to intent accuracy. Data is divided between utterance reordering and third utterance generation.}
\centering
\label{fig:reorder-proportion}
\end{figure}

\paragraph{Sentence reordering importance}
We studied the impact of the additional semi-supervised task (see \S~\ref{sec:training-regimes}) on intent prediction. 
Notably, creating a semi-supervised dataset for the utterance reordering task is straightforward. Thus, this task can be added to any model, using any dialogue dataset. 

Figure \ref{fig:reorder-proportion} shows intent-classification accuracy when training with five utterance reordering ratio alternatives ($0.0, 0.1, 0.3, 0.5, 1.0$) on the \textit{MulitWOZ} and \textit{SGD} datasets.
Per the $0.0$ alternative, we used all data for the utterance generation task, while per the $1.0$ alternative, we used all data for the utterance reordering task.
The results indicate similar trends in both datasets, showing that using a small portion ($0.1$) of utterance reordering data is most beneficial. 
The proportion parameter affects model performance, with a gap higher than $1.1\%$ between best- and worst-performing model in both datasets. 

\begin{table}
\small
\centering
\begin{tabular}{c | c | c | c | c | c}
     & \multicolumn{2}{c |}{\emph{3-gen}} & \multicolumn{2}{c|}{\emph{3-5xg}} & \emph{3-rnd}\\
    
     \textbf{\textit{MultiWOZ}} & \emph{PART} & \emph{ALL} & \emph{PART} & \emph{ALL} & --\\
     \hline
     \emph{SDC}       & 94.4 & 95.5 & 95.7 & \textbf{96.6} & 92.7\\
     \emph{PART-SDC}  & 95.3 & 95.3 & 96.6 & \textbf{97.0} & 89.7\\
     \emph{ALL}    & 96.6 & \textbf{97.9} & 97.4 & \textbf{97.9}& 92.3\\
     \emph{ALL-SDC} & 96.6 & 96.6 & \textbf{98.3} & \textbf{98.3} & 94.8\\
     \toprule
     \textbf{\textit{SGD}} & \emph{PART} & \emph{ALL} & \emph{PART} & \emph{ALL} & --\\
     \hline
     \emph{SDC}       & 72.3 & 72.6 & 73.4 & \textbf{74.0} & 62.3\\
     \emph{PART-SDC}  & 73.1 & 73.1 & 73.7 & \textbf{74.3} & 64.7\\
     \emph{ALL}    &  75.9 & 75.3 & 76.7 & \textbf{77.2} & 65.8\\
     \emph{ALL-SDC} & 73.7 & 72.6 & 75.6 & \textbf{75.8} & 64.3\\
     \toprule
     \textbf{\textit{EDU}} & \emph{PART} & \emph{ALL} & \emph{PART} & \emph{ALL} & --\\
     \hline
     \emph{SDC}    & $+2.7$ & $+4.6$ & $+3.7$ & $\textbf{+5.5}$ & $+2.7$\\
     \emph{PART-SDC}  & $+1.8$ & $+4.6$ & $\textbf{+5.5}$ & $\textbf{+5.5}$ & $-2.7$\\
     \emph{ALL}    & $+3.7$ & $\textbf{+5.5}$ & $+3.7$ & $+3.7$ & $+1.8$\\
     \emph{ALL-SDC} & $+7.2$ & $\textbf{+9.0}$ & $+7.2$ & $+7.2$ & $+6.3$\\

     
\end{tabular}
\caption{Comparison of quality of sentences generated by \emph{PART} and \emph{ALL} models, and sampled with three different approaches: once (\emph{3-gen}); five-times with majority voting (\emph{3-5xg}); and without context (\emph{3-rnd})}
\label{tab:comp-generation-lookahead}
\end{table}

\paragraph{Look-ahead for intent prediction}
Even though our main table reports results from majority voting, this was not our first approach to generation-for-look-ahead.
Naively, we started our experiments with generating just a single look-ahead utterance.
We report the results from this experiment in the \emph{3-gen} columns of Table \ref{tab:comp-generation-lookahead}.
When this approach failed to meet the expected performance, we turned to sample several (e.g., five) times from the generative model and used majority voting to elect the predicted intent.
Results of this experiment appear in the  \emph{3-5xg} columns of Table \ref{tab:comp-generation-lookahead}.
Notably, majority voting improves performance across most models and datasets.
As a last baseline-step of our analysis, we used random generated utterances for look-ahead.
The results appear in the \emph{3-rnd} column of Table \ref{tab:comp-generation-lookahead}.
Clearly, the results degrade significantly across all datasets and all models, reconfirming the importance of qualitative third utterances for accurate intent prediction.

\paragraph{Qualitative look-ahead utterances}
To analyse the ability of the model to generate qualitative look-ahead utterances, we considered two model alternatives: \textit{PART}, which was trained on all tasks but not on the supervised intent prediction, and \emph{ALL}, which was trained on all tasks, intent-prediction included.
We used all the above-mentioned \emph{SDC} model variants to evaluate utterance quality, as indicated by the rows of Table \ref{tab:comp-generation-lookahead}.
Each of these models was tested with utterances generated by both generation model alternatives (\emph{PART} and \emph{ALL} columns of Table \ref{tab:comp-generation-lookahead}) and under both look-ahead techniques (\emph{3-gen} and \emph{3-5xg}).

Table \ref{tab:comp-generation-lookahead} shows that \emph{ALL} is a statistically stronger generative-model, outperforing PART in 15 out of 24 look-ahead experiments.
In comparison, only in 2 out of the 24 experiments did the models under evaluation achieve higher accuracy when using utterances generated by the \textit{PART} model.
For both generative models, generating five user utterances and using majority voting to choose among them outperforms the single-generation alternative.
This result is consistent with results presented in \S~\ref{sec:results}, and confirms that the quality of generated utterances does not depend on the quality of the discriminative model used for intent prediction.

\begin{figure}
\includegraphics[scale=0.34]{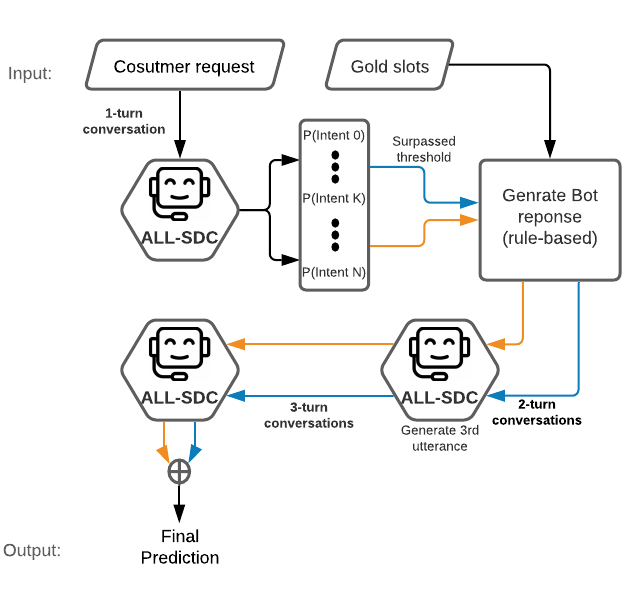}
\centering
\caption{Using \textit{ALL-SDC} for solving intent conflicts with counterfactual based algorithm. The model assigns a probability for each of the intents given an input, i.e. a 1-turn conversation (costumer request). Then, if more than one intent surpasses a pre-chosen threshold bound, we choose the top two intents (orange and blue arrows) and generate a counterfactual conversation for each. Finally, the model predicts the intent given each of the conversation and ensembles both predictions to choose a final intent.}
\centering
\label{fig:counterfactual}
\end{figure}

\paragraph{Solving intent conflicts with counterfactual user utterances}
Most goal oriented dialogue systems use specific approaches for solving ambiguity in intent prediction. For example, the bot can present the user with the conflicting intents and let her choose the right one, or it can ask a clarification question and wait for user response.
Here, we suggest a novel counterfactual approach for solving intent ambiguity.
In contrast to other approaches, ours doesn't require any additional input from the user.
To achieve that we use the \textit{3UG} model to generate counterfactual user utterance for each of the conflicting intents (see Figure~\ref{fig:counterfactual}).
We test our approach with the following setup:
(1) Fit an \textit{ALL-SDC} model with an intent classification head to the SGD training set and use it to predict intent of test examples with single user-utterance;
(2) Identify conflicting intents by setting a threshold bound on the last softmax layer of the \textit{ALL-SDC} classifier;
(3) For each conflicting intent, generate the bot’s response (second utterance). Since we do not able to generate bot responses in our experimental setup , we mimic the generation process by choosing a response utterance that is associated with the predicted intent and has the most overlapping slots with the gold response (gold slots are provided in the SGD dataset);
(4) Generate the third utterance by applying the \textit{3UG} model to the first two utterances;
(5) Predict the intent based on the ensemble of predictions for all look-ahead conversations, where each conversation now has three successive utterances.

While, ideally, one may detect conflicting intents by applying a threshold bound (\textit{Threshold-based conflicts}), we also examine our counterfactual algorithm when given a hindsight regards conflicting intents.
We examine two types of hindsight:
(1) Identifying all mistaken examples and count the two top predicted intents as the conflict intents (\textit{Mistake Oracle}); and 
(2) Identifying mistaken examples in which the second top predicted intent is the gold intent (\textit{Conflict-oracle}).  

In Table~\ref{tab:counterfactual-conflicts}, we report experiment results.
As seen, with the \textit{Conflict-oracle} hindsight the algorithm fixes more than $30\%$ of the mistakes examples. 
This illustrates the potential of our algorithm.
To fully realize this potential one must have an intent classifier with high accuracy/confidence correlation.
However, when using our classifier's confidence, algorithm achieves modest improvement.
This can be attributed to the un-calibrated nature of the used classifier and its failure, in most cases, to predict the gold intent as the second- or third--top prediction.

\begin{table}
\small
\centering
\begin{tabular}{l | c }

     \textbf{Conflicts Selection Method} & \textbf{Error reduction}  \\
     \hline
     Conflict-oracle & $31\%$ \\
     Mistake-oracle &  $ 8\%$ \\
     Threshold-based conflicts  & $2\%$  \\

\end{tabular}
\caption{Error reduction for our counterfactual-based conflict solving algorithm.}
\label{tab:counterfactual-conflicts}
\end{table}

\section{Conclusions}
\label{sec:conclusions}
In this work we show how recent text-to-text models can benefit from abundant past conversation data to improve the intent prediction along the course of the dialogue.
Specifically, we show that training text-to-text models to generate successive user utterances as a secondary task in a multi-task training regime improves the performance of the main intent prediction task. 
We further show that the real power of these generative text-to-text models, trained on vast amounts of unlabeled data, lies in their ability to synthetically sample possible user responses and use them as a look ahead signal to improve intent prediction in inference time.

\bibliography{tacl2018}

\begin{thebibliography}{46}
\expandafter\ifx\csname natexlab\endcsname\relax\def\natexlab#1{#1}\fi

\bibitem[{Anaby{-}Tavor et~al.(2020)Anaby{-}Tavor, Carmeli, Goldbraich, Kantor,
  Kour, Shlomov, Tepper, and Zwerdling}]{DBLP:conf/aaai/Anaby-TavorCGKK20}
Ateret Anaby{-}Tavor, Boaz Carmeli, Esther Goldbraich, Amir Kantor, George
  Kour, Segev Shlomov, Naama Tepper, and Naama Zwerdling. 2020.
\newblock \href {https://aaai.org/ojs/index.php/AAAI/article/view/6233} {Do not
  have enough data? deep learning to the rescue!}
\newblock In \emph{The Thirty-Fourth {AAAI} Conference on Artificial
  Intelligence, {AAAI} 2020, The Thirty-Second Innovative Applications of
  Artificial Intelligence Conference, {IAAI} 2020, The Tenth {AAAI} Symposium
  on Educational Advances in Artificial Intelligence, {EAAI} 2020, New York,
  NY, USA, February 7-12, 2020}, pages 7383--7390. {AAAI} Press.

\bibitem[{Asri et~al.(2016)Asri, He, and
  Suleman}]{DBLP:conf/interspeech/AsriHS16}
Layla~El Asri, Jing He, and Kaheer Suleman. 2016.
\newblock \href {https://doi.org/10.21437/Interspeech.2016-1175} {A
  sequence-to-sequence model for user simulation in spoken dialogue systems}.
\newblock In \emph{Interspeech 2016, 17th Annual Conference of the
  International Speech Communication Association, San Francisco, CA, USA,
  September 8-12, 2016}, pages 1151--1155. {ISCA}.

\bibitem[{Bocklisch et~al.(2017)Bocklisch, Faulkner, Pawlowski, and
  Nichol}]{bocklisch2017rasa:27}
Tom Bocklisch, Joey Faulkner, Nick Pawlowski, and Alan Nichol. 2017.
\newblock Rasa: Open source language understanding and dialogue management.
\newblock \emph{arXiv preprint arXiv:1712.05181}.

\bibitem[{Budzianowski et~al.(2018)Budzianowski, Wen, Tseng, Casanueva, Ultes,
  Ramadan, and Ga{\v{s}}i{\'c}}]{budzianowski2018multiwoz:2}
Pawe{\l} Budzianowski, Tsung-Hsien Wen, Bo-Hsiang Tseng, Inigo Casanueva,
  Stefan Ultes, Osman Ramadan, and Milica Ga{\v{s}}i{\'c}. 2018.
\newblock Multiwoz-a large-scale multi-domain wizard-of-oz dataset for
  task-oriented dialogue modelling.
\newblock \emph{arXiv preprint arXiv:1810.00278}.

\bibitem[{Chen et~al.(2017)Chen, Liu, Yin, and Tang}]{chen2017survey:41}
Hongshen Chen, Xiaorui Liu, Dawei Yin, and Jiliang Tang. 2017.
\newblock A survey on dialogue systems: Recent advances and new frontiers.
\newblock \emph{Acm Sigkdd Explorations Newsletter}, 19(2):25--35.

\bibitem[{Chen et~al.(2019)Chen, Zhuo, and
  Wang}]{DBLP:journals/corr/abs-1902-10909}
Qian Chen, Zhu Zhuo, and Wen Wang. 2019.
\newblock \href {http://arxiv.org/abs/1902.10909} {{BERT} for joint intent
  classification and slot filling}.
\newblock \emph{CoRR}, abs/1902.10909.

\bibitem[{Chen et~al.(2016)Chen, Hakkani{-}T{\"{u}}r, T{\"{u}}r, Gao, and
  Deng}]{DBLP:conf/interspeech/ChenHTGD16}
Yun{-}Nung Chen, Dilek Hakkani{-}T{\"{u}}r, G{\"{o}}khan T{\"{u}}r, Jianfeng
  Gao, and Li~Deng. 2016.
\newblock \href {https://doi.org/10.21437/Interspeech.2016-312} {End-to-end
  memory networks with knowledge carryover for multi-turn spoken language
  understanding}.
\newblock In \emph{Interspeech 2016, 17th Annual Conference of the
  International Speech Communication Association, San Francisco, CA, USA,
  September 8-12, 2016}, pages 3245--3249. {ISCA}.

\bibitem[{Cheng et~al.(2020)Cheng, Agrawal, Alonso, Bhargava, Driesen, Flego,
  Kaplan, Kartsaklis, Li, Piraviperumal et~al.}]{cheng2020conversational:50}
Jianpeng Cheng, Devang Agrawal, Hector~Martinez Alonso, Shruti Bhargava, Joris
  Driesen, Federico Flego, Dain Kaplan, Dimitri Kartsaklis, Lin Li, Dhivya
  Piraviperumal, et~al. 2020.
\newblock Conversational semantic parsing for dialog state tracking.
\newblock \emph{arXiv preprint arXiv:2010.12770}.

\bibitem[{Devlin et~al.(2019)Devlin, Chang, Lee, and
  Toutanova}]{devlin2018bert}
Jacob Devlin, Ming{-}Wei Chang, Kenton Lee, and Kristina Toutanova. 2019.
\newblock \href {https://doi.org/10.18653/v1/n19-1423} {{BERT:} pre-training of
  deep bidirectional transformers for language understanding}.
\newblock In \emph{Proceedings of the 2019 Conference of the North American
  Chapter of the Association for Computational Linguistics: Human Language
  Technologies, {NAACL-HLT} 2019, Minneapolis, MN, USA, June 2-7, 2019, Volume
  1 (Long and Short Papers)}, pages 4171--4186. Association for Computational
  Linguistics.

\bibitem[{Feng et~al.(2020)Feng, Ren, Chen, Sun, Li, and
  Sun}]{feng2020regularizing:51}
Shaoxiong Feng, Xuancheng Ren, Hongshen Chen, Bin Sun, Kan Li, and Xu~Sun.
  2020.
\newblock Regularizing dialogue generation by imitating implicit scenarios.
\newblock \emph{arXiv preprint arXiv:2010.01893}.

\bibitem[{Gao et~al.(2018)Gao, Galley, and Li}]{gao2018neural:16}
Jianfeng Gao, Michel Galley, and Lihong Li. 2018.
\newblock Neural approaches to conversational ai.
\newblock In \emph{The 41st International ACM SIGIR Conference on Research \&
  Development in Information Retrieval}, pages 1371--1374.

\bibitem[{Goo et~al.(2018)Goo, Gao, Hsu, Huo, Chen, Hsu, and
  Chen}]{DBLP:conf/naacl/GooGHHCHC18}
Chih{-}Wen Goo, Guang Gao, Yun{-}Kai Hsu, Chih{-}Li Huo, Tsung{-}Chieh Chen,
  Keng{-}Wei Hsu, and Yun{-}Nung Chen. 2018.
\newblock \href {https://doi.org/10.18653/v1/n18-2118} {Slot-gated modeling for
  joint slot filling and intent prediction}.
\newblock In \emph{Proceedings of the 2018 Conference of the North American
  Chapter of the Association for Computational Linguistics: Human Language
  Technologies, NAACL-HLT, New Orleans, Louisiana, USA, June 1-6, 2018, Volume
  2 (Short Papers)}, pages 753--757. Association for Computational Linguistics.

\bibitem[{Hancock et~al.(2019)Hancock, Bordes, Mazare, and
  Weston}]{hancock2019learning:6}
Braden Hancock, Antoine Bordes, Pierre-Emmanuel Mazare, and Jason Weston. 2019.
\newblock Learning from dialogue after deployment: Feed yourself, chatbot!
\newblock \emph{arXiv preprint arXiv:1901.05415}.

\bibitem[{He et~al.(2018)He, Chen, Balakrishnan, and
  Liang}]{DBLP:conf/emnlp/HeCBL18}
He~He, Derek Chen, Anusha Balakrishnan, and Percy Liang. 2018.
\newblock \href {https://doi.org/10.18653/v1/d18-1256} {Decoupling strategy and
  generation in negotiation dialogues}.
\newblock In \emph{Proceedings of the 2018 Conference on Empirical Methods in
  Natural Language Processing, Brussels, Belgium, October 31 - November 4,
  2018}, pages 2333--2343. Association for Computational Linguistics.

\bibitem[{Kobsa(1990)}]{DBLP:journals/ais/Kobsa90}
Alfred Kobsa. 1990.
\newblock \href {https://doi.org/10.1007/BF01889941} {User modeling in dialog
  systems: Potentials and hazards}.
\newblock \emph{{AI} Soc.}, 4(3):214--231.

\bibitem[{Kobsa and Carberry(1989)}]{DBLP:books/lib/KobsaC89}
Alfred Kobsa and Sandra Carberry. 1989.
\newblock \href {https://www.worldcat.org/oclc/18463611} {\emph{User models in
  dialog systems}}.
\newblock Symbolic computation : Artificial intelligence. Springer.

\bibitem[{Kreyssig et~al.(2018)Kreyssig, Casanueva, Budzianowski, and
  Gasic}]{DBLP:conf/sigdial/KreyssigCBG18}
Florian Kreyssig, I{\~{n}}igo Casanueva, Pawel Budzianowski, and Milica Gasic.
  2018.
\newblock \href {https://doi.org/10.18653/v1/w18-5007} {Neural user simulation
  for corpus-based policy optimisation of spoken dialogue systems}.
\newblock In \emph{Proceedings of the 19th Annual SIGdial Meeting on Discourse
  and Dialogue, Melbourne, Australia, July 12-14, 2018}, pages 60--69.
  Association for Computational Linguistics.

\bibitem[{Li et~al.(2016{\natexlab{a}})Li, Galley, Brockett, Spithourakis, Gao,
  and Dolan}]{DBLP:conf/acl/LiGBSGD16}
Jiwei Li, Michel Galley, Chris Brockett, Georgios~P. Spithourakis, Jianfeng
  Gao, and William~B. Dolan. 2016{\natexlab{a}}.
\newblock \href {https://doi.org/10.18653/v1/p16-1094} {A persona-based neural
  conversation model}.
\newblock In \emph{Proceedings of the 54th Annual Meeting of the Association
  for Computational Linguistics, {ACL} 2016, August 7-12, 2016, Berlin,
  Germany, Volume 1: Long Papers}. The Association for Computer Linguistics.

\bibitem[{Li et~al.(2016{\natexlab{b}})Li, Lipton, Dhingra, Li, Gao, and
  Chen}]{DBLP:journals/corr/LiLDLGC16}
Xiujun Li, Zachary~C. Lipton, Bhuwan Dhingra, Lihong Li, Jianfeng Gao, and
  Yun{-}Nung Chen. 2016{\natexlab{b}}.
\newblock \href {http://arxiv.org/abs/1612.05688} {A user simulator for
  task-completion dialogues}.
\newblock \emph{CoRR}, abs/1612.05688.

\bibitem[{Lin and Walker(2011)}]{DBLP:conf/aiide/LinW11}
Grace~I. Lin and Marilyn~A. Walker. 2011.
\newblock \href
  {http://www.aaai.org/ocs/index.php/AIIDE/AIIDE11/paper/view/4065} {All the
  world's a stage: Learning character models from film}.
\newblock In \emph{Proceedings of the Seventh {AAAI} Conference on Artificial
  Intelligence and Interactive Digital Entertainment, {AIIDE} 2011, October
  10-14, 2011, Stanford, California, {USA}}. The {AAAI} Press.

\bibitem[{Liu and Lane(2016)}]{DBLP:conf/interspeech/LiuL16}
Bing Liu and Ian Lane. 2016.
\newblock \href {https://doi.org/10.21437/Interspeech.2016-1352}
  {Attention-based recurrent neural network models for joint intent detection
  and slot filling}.
\newblock In \emph{Interspeech 2016, 17th Annual Conference of the
  International Speech Communication Association, San Francisco, CA, USA,
  September 8-12, 2016}, pages 685--689. {ISCA}.

\bibitem[{Liu and Lane(2017)}]{DBLP:conf/interspeech/LiuL17}
Bing Liu and Ian Lane. 2017.
\newblock \href
  {http://www.isca-speech.org/archive/Interspeech\_2017/abstracts/1326.html}
  {An end-to-end trainable neural network model with belief tracking for
  task-oriented dialog}.
\newblock In \emph{Interspeech 2017, 18th Annual Conference of the
  International Speech Communication Association, Stockholm, Sweden, August
  20-24, 2017}, pages 2506--2510. {ISCA}.

\bibitem[{Liu et~al.(2019)Liu, Ott, Goyal, Du, Joshi, Chen, Levy, Lewis,
  Zettlemoyer, and Stoyanov}]{liu2019roberta}
Yinhan Liu, Myle Ott, Naman Goyal, Jingfei Du, Mandar Joshi, Danqi Chen, Omer
  Levy, Mike Lewis, Luke Zettlemoyer, and Veselin Stoyanov. 2019.
\newblock \href {http://arxiv.org/abs/1907.11692} {Roberta: {A} robustly
  optimized {BERT} pretraining approach}.
\newblock \emph{CoRR}, abs/1907.11692.

\bibitem[{Luo et~al.(2014)Luo, Zhang, and Yang}]{luo2014win:45}
Jiyun Luo, Sicong Zhang, and Hui Yang. 2014.
\newblock Win-win search: Dual-agent stochastic game in session search.
\newblock In \emph{Proceedings of the 37th international ACM SIGIR conference
  on Research \& development in information retrieval}, pages 587--596.

\bibitem[{Ma et~al.(2019)Ma, Zeng, Zhu, Li, Yang, Yao, Zhou, and
  Shen}]{DBLP:journals/corr/abs-1912-09297}
Yue Ma, Zengfeng Zeng, Dawei Zhu, Xuan Li, Yiying Yang, Xiaoyuan Yao, Kaijie
  Zhou, and Jianping Shen. 2019.
\newblock \href {http://arxiv.org/abs/1912.09297} {An end-to-end dialogue state
  tracking system with machine reading comprehension and wide {\&} deep
  classification}.
\newblock \emph{CoRR}, abs/1912.09297.

\bibitem[{Martin et~al.(2020)Martin, M{\"{u}}ller, Su{\'{a}}rez, Dupont,
  Romary, de~la Clergerie, Seddah, and Sagot}]{DBLP:conf/acl/MartinMSDRCSS20}
Louis Martin, Benjamin M{\"{u}}ller, Pedro Javier~Ortiz Su{\'{a}}rez, Yoann
  Dupont, Laurent Romary, {\'{E}}ric de~la Clergerie, Djam{\'{e}} Seddah, and
  Beno{\^{\i}}t Sagot. 2020.
\newblock \href {https://www.aclweb.org/anthology/2020.acl-main.645/}
  {Camembert: a tasty french language model}.
\newblock In \emph{Proceedings of the 58th Annual Meeting of the Association
  for Computational Linguistics, {ACL} 2020, Online, July 5-10, 2020}, pages
  7203--7219. Association for Computational Linguistics.

\bibitem[{Nouri and Hosseini{-}Asl(2018)}]{DBLP:journals/corr/abs-1812-00899}
Elnaz Nouri and Ehsan Hosseini{-}Asl. 2018.
\newblock \href {http://arxiv.org/abs/1812.00899} {Toward scalable neural
  dialogue state tracking model}.
\newblock \emph{CoRR}, abs/1812.00899.

\bibitem[{Radford et~al.(2019)Radford, Wu, Child, Luan, Amodei, and
  Sutskever}]{radford2019language}
Alec Radford, Jeffrey Wu, Rewon Child, David Luan, Dario Amodei, and Ilya
  Sutskever. 2019.
\newblock \href
  {https://www.ceid.upatras.gr/webpages/faculty/zaro/teaching/alg-ds/PRESENTATIONS/PAPERS/2019-Radford-et-al_Language-Models-Are-Unsupervised-Multitask-%20Learners.pdf}
  {Language models are unsupervised multitask learners}.
\newblock \emph{OpenAI Blog}, 1(8).

\bibitem[{Raffel et~al.(2019)Raffel, Shazeer, Roberts, Lee, Narang, Matena,
  Zhou, Li, and Liu}]{DBLP:journals/corr/abs-1910-10683}
Colin Raffel, Noam Shazeer, Adam Roberts, Katherine Lee, Sharan Narang, Michael
  Matena, Yanqi Zhou, Wei Li, and Peter~J. Liu. 2019.
\newblock Exploring the limits of transfer learning with a unified text-to-text
  transformer.
\newblock \emph{CoRR}, abs/1910.10683.

\bibitem[{Ralston et~al.(2019)Ralston, Chen, Isah, and
  Zulkernine}]{ralston2019voice:46}
Kennedy Ralston, Yuhao Chen, Haruna Isah, and Farhana Zulkernine. 2019.
\newblock A voice interactive multilingual student support system using ibm
  watson.
\newblock In \emph{2019 18th IEEE International Conference On Machine Learning
  And Applications (ICMLA)}, pages 1924--1929. IEEE.

\bibitem[{Rastogi et~al.(2019)Rastogi, Zang, Sunkara, Gupta, and
  Khaitan}]{rastogi2019towards:9}
Abhinav Rastogi, Xiaoxue Zang, Srinivas Sunkara, Raghav Gupta, and Pranav
  Khaitan. 2019.
\newblock Towards scalable multi-domain conversational agents: The
  schema-guided dialogue dataset.
\newblock \emph{arXiv preprint arXiv:1909.05855}.

\bibitem[{Rastogi et~al.(2020)Rastogi, Zang, Sunkara, Gupta, and
  Khaitan}]{DBLP:conf/aaai/RastogiZSGK20}
Abhinav Rastogi, Xiaoxue Zang, Srinivas Sunkara, Raghav Gupta, and Pranav
  Khaitan. 2020.
\newblock \href {https://aaai.org/ojs/index.php/AAAI/article/view/6394}
  {Towards scalable multi-domain conversational agents: The schema-guided
  dialogue dataset}.
\newblock In \emph{The Thirty-Fourth {AAAI} Conference on Artificial
  Intelligence, {AAAI} 2020, The Thirty-Second Innovative Applications of
  Artificial Intelligence Conference, {IAAI} 2020, The Tenth {AAAI} Symposium
  on Educational Advances in Artificial Intelligence, {EAAI} 2020, New York,
  NY, USA, February 7-12, 2020}, pages 8689--8696. {AAAI} Press.

\bibitem[{Ravuri and Stolcke(2015)}]{DBLP:conf/interspeech/RavuriS15}
Suman~V. Ravuri and Andreas Stolcke. 2015.
\newblock \href
  {http://www.isca-speech.org/archive/interspeech\_2015/i15\_0135.html}
  {Recurrent neural network and {LSTM} models for lexical utterance
  classification}.
\newblock In \emph{{INTERSPEECH} 2015, 16th Annual Conference of the
  International Speech Communication Association, Dresden, Germany, September
  6-10, 2015}, pages 135--139. {ISCA}.

\bibitem[{Rothe et~al.(2020)Rothe, Narayan, and
  Severyn}]{DBLP:journals/tacl/RotheNS20}
Sascha Rothe, Shashi Narayan, and Aliaksei Severyn. 2020.
\newblock \href {https://transacl.org/ojs/index.php/tacl/article/view/1849}
  {Leveraging pre-trained checkpoints for sequence generation tasks}.
\newblock \emph{Trans. Assoc. Comput. Linguistics}, 8:264--280.

\bibitem[{Sanh et~al.(2019)Sanh, Debut, Chaumond, and
  Wolf}]{DBLP:journals/corr/abs-1910-01108}
Victor Sanh, Lysandre Debut, Julien Chaumond, and Thomas Wolf. 2019.
\newblock \href {http://arxiv.org/abs/1910.01108} {Distilbert, a distilled
  version of {BERT:} smaller, faster, cheaper and lighter}.
\newblock \emph{CoRR}, abs/1910.01108.

\bibitem[{Schatzmann et~al.(2007)Schatzmann, Thomson, Weilhammer, Ye, and
  Young}]{DBLP:conf/naacl/SchatzmannTWYY07}
Jost Schatzmann, Blaise Thomson, Karl Weilhammer, Hui Ye, and Steve~J. Young.
  2007.
\newblock \href {https://www.aclweb.org/anthology/N07-2038/} {Agenda-based user
  simulation for bootstrapping a {POMDP} dialogue system}.
\newblock In \emph{Human Language Technology Conference of the North American
  Chapter of the Association of Computational Linguistics, Proceedings, April
  22-27, 2007, Rochester, New York, {USA}}, pages 149--152. The Association for
  Computational Linguistics.

\bibitem[{Schatzmann et~al.(2006)Schatzmann, Weilhammer, Stuttle, and
  Young}]{DBLP:journals/ker/SchatzmannWSY06}
Jost Schatzmann, Karl Weilhammer, Matthew~N. Stuttle, and Steve~J. Young. 2006.
\newblock \href {https://doi.org/10.1017/S0269888906000944} {A survey of
  statistical user simulation techniques for reinforcement-learning of dialogue
  management strategies}.
\newblock \emph{Knowl. Eng. Rev.}, 21(2):97--126.

\bibitem[{Shin et~al.(2019)Shin, Xu, Madotto, and Fung}]{shin2019happybot:48}
Jamin Shin, Peng Xu, Andrea Madotto, and Pascale Fung. 2019.
\newblock Happybot: Generating empathetic dialogue responses by improving user
  experience look-ahead.
\newblock \emph{arXiv preprint arXiv:1906.08487}.

\bibitem[{Tur et~al.(2012)Tur, Deng, Hakkani-T{\"u}r, and
  He}]{tur2012towards:43}
Gokhan Tur, Li~Deng, Dilek Hakkani-T{\"u}r, and Xiaodong He. 2012.
\newblock Towards deeper understanding: Deep convex networks for semantic
  utterance classification.
\newblock In \emph{2012 IEEE international conference on acoustics, speech and
  signal processing (ICASSP)}, pages 5045--5048. IEEE.

\bibitem[{Vaswani et~al.(2017)Vaswani, Shazeer, Parmar, Uszkoreit, Jones,
  Gomez, Kaiser, and Polosukhin}]{DBLP:conf/nips/VaswaniSPUJGKP17}
Ashish Vaswani, Noam Shazeer, Niki Parmar, Jakob Uszkoreit, Llion Jones,
  Aidan~N. Gomez, Lukasz Kaiser, and Illia Polosukhin. 2017.
\newblock \href {http://papers.nips.cc/paper/7181-attention-is-all-you-need}
  {Attention is all you need}.
\newblock In \emph{Advances in Neural Information Processing Systems 30: Annual
  Conference on Neural Information Processing Systems 2017, 4-9 December 2017,
  Long Beach, CA, {USA}}, pages 5998--6008.

\bibitem[{Weston(2016)}]{weston2016dialog:13}
Jason~E Weston. 2016.
\newblock Dialog-based language learning.
\newblock In \emph{Advances in Neural Information Processing Systems}, pages
  829--837.

\bibitem[{Xu and Sarikaya(2013)}]{DBLP:conf/asru/XuS13}
Puyang Xu and Ruhi Sarikaya. 2013.
\newblock \href {https://doi.org/10.1109/ASRU.2013.6707709} {Convolutional
  neural network based triangular {CRF} for joint intent detection and slot
  filling}.
\newblock In \emph{2013 {IEEE} Workshop on Automatic Speech Recognition and
  Understanding, Olomouc, Czech Republic, December 8-12, 2013}, pages 78--83.
  {IEEE}.

\bibitem[{Yan et~al.(2017)Yan, Duan, Chen, Zhou, Zhou, and
  Li}]{yan2017building:44}
Zhao Yan, Nan Duan, Peng Chen, Ming Zhou, Jianshe Zhou, and Zhoujun Li. 2017.
\newblock Building task-oriented dialogue systems for online shopping.
\newblock In \emph{Proceedings of the Thirty-First AAAI Conference on
  Artificial Intelligence}, pages 4618--4625.

\bibitem[{Yang et~al.(2019)Yang, Dai, Yang, Carbonell, Salakhutdinov, and
  Le}]{yang2019xlnet}
Zhilin Yang, Zihang Dai, Yiming Yang, Jaime~G. Carbonell, Ruslan Salakhutdinov,
  and Quoc~V. Le. 2019.
\newblock \href
  {http://papers.nips.cc/paper/8812-xlnet-generalized-autoregressive-pretraining-for-language-understanding}
  {Xlnet: Generalized autoregressive pretraining for language understanding}.
\newblock In \emph{Advances in Neural Information Processing Systems 32: Annual
  Conference on Neural Information Processing Systems 2019, NeurIPS 2019, 8-14
  December 2019, Vancouver, BC, Canada}, pages 5754--5764.

\bibitem[{Zhang et~al.(2015)Zhang, Zhao, and LeCun}]{DBLP:conf/nips/ZhangZL15}
Xiang Zhang, Junbo~Jake Zhao, and Yann LeCun. 2015.
\newblock \href
  {http://papers.nips.cc/paper/5782-character-level-convolutional-networks-for-text-classification}
  {Character-level convolutional networks for text classification}.
\newblock In \emph{Advances in Neural Information Processing Systems 28: Annual
  Conference on Neural Information Processing Systems 2015, December 7-12,
  2015, Montreal, Quebec, Canada}, pages 649--657.

\bibitem[{Zhao and Wu(2016)}]{DBLP:conf/interspeech/ZhaoW16}
Zhiwei Zhao and Youzheng Wu. 2016.
\newblock \href {https://doi.org/10.21437/Interspeech.2016-354}
  {Attention-based convolutional neural networks for sentence classification}.
\newblock In \emph{Interspeech 2016, 17th Annual Conference of the
  International Speech Communication Association, San Francisco, CA, USA,
  September 8-12, 2016}, pages 705--709. {ISCA}.

\end{thebibliography}
\bibliographystyle{acl_natbib}

\appendix
\section{Auxiliary Tasks}
\label{sec:auxiliary-tasks}
We took further advantage of the T5 model's ability to share parameters across tasks and defined several auxiliary dialogue-related tasks:\\
\textbf{Utterance reordering:} shuffle dialogue utterances and train the model to correctly reorder them.\\
\textbf{Conversation escalation:} predict \textit{false} if the bot contains the entire conversation and \textit{true} if it was eventually handled by a human agent.\\
\textbf{Utterance repetition:} predict \textit{true} for repetitive bot utterances or rephrased user utterances, and \textit{false} otherwise.\\
We used various weak labeling techniques to add labels to the raw dialogue data for each of these tasks.
We then used these semi-supervised datasets together with the supervised full-dialogue dataset and the unsupervised dialogue data to train a T5 model to predict the user intent.
We evaluated the performance gained from each of these datasets by introducing them to the T5 model in different orders and amount ratios, and measured the intent prediction accuracy.

We performed an analysis by examining the performance of additional auxiliary tasks.  
Table \ref{tab:auxiliary importance} presents results for the \textit{EDU} and \textit{SGD} datasets.
For the \textit{EDU} dataset, we considered all auxiliary tasks presented in \S~\ref{sec:auxiliary-tasks}.
For \textit{SGD}, we did not include \textit{utterance-repetition} and \textit{conversation-escalation} as this dataset does not contain these phenomena.
We report intent prediction results achieved with a one-step training regime, similar to the training regime that we use to train the \emph{ALL} model.

Clearly, combining all tasks together performs better than using each task separately.
The gap between the \emph{ALL} model and the best-performing single-task model is $1.7\%$ for \textit{EDU} and $0.4\%$ for \textit{SGD}.
Furthermore, when used separately, the \textit{escalation} and \textit{repetitious} tasks perform sub-optimally, with measured accuracy of $+1.8\%$ and $0.0\%$ respectively, compared to $+4.6\%$ and $+3.6\%$ achieved by the utterance generation and reordering tasks, respectively.
This did not come as surprise, since these signals correlate with dialogue outcome more than with a specific user intent. 

\begin{table}
\small
\centering
\begin{tabular}{l | c | c }
    & \multicolumn{1}{c |}{\textbf{\textit{EDU}}} & \multicolumn{1}{c}{\textbf{\textit{SGD}}} \\
    & \multicolumn{1}{c |}{\textbf{\emph{3-u}}} & \multicolumn{1}{c}{\textbf{\emph{3-u}}} \\
    \hline
     Utterance Generation  & $+4.6$ & 76.2 \\ 
     Reordering  & $+3.6$  & 75.8 \\ 
     Escalation  & $+1.8$  & - \\
     Repetitious  & $+0.0$  & - \\
        \hline
     \emph{ALL}      & $+6.3$ & 76.6 \\
\end{tabular}
\caption{Model performance using various auxiliary tasks.}
\label{tab:auxiliary importance}
\end{table}

\section{Alternative Classification Models}
\label{sec:BERT}
In this analysis, we focus on a potential alternative
modeling solution.
As described in the main paper, we direct our modeling efforts at T5 variants \citep{DBLP:journals/corr/abs-1910-10683}: through baselines and our own contributions.  However, another possible modeling direction for the task could use a state-of-the-art discriminative classifier \citep{devlin2018bert,yang2019xlnet,liu2019roberta}. 

Thus, as another baseline, we used a BERT classifier \citep{devlin2018bert}, which we trained to predict the intent from three successive utterances, denoted as \emph{SDC-BERT}.
In Table~\ref{tab:comparing bert performance to t5.}, we report models' accuracy across the three different datasets: \textit{Multiwoz}, \textit{SGD}, and \textit{EDU}.
We further tested this model under our look-ahead scenario, \emph{3-gen}, while using our trained T5 \emph{ALL} model to generate the look-ahead utterances.
For a complete picture, we present the results of our three T5 variants, \emph{SDC}, \emph{ALL}, and \emph{ALL-SDC}, even though these results have already been presented in the main paper.
 
Our unsupervised and semi-supervised data-integration methods improved SDC performance across  most  training  regimes; \emph{ALL} and \emph{ALL-SDC} outperform \emph{SDC-BERT} in five out of six different setups.
The performance of the T5 \emph{SDC} and \emph{SDC-BERT} models that use only supervised data, are on a par. In three out of six setups, the T5-based classifier outperforms BERT, and vice versa.
However, looking deeper, \emph{SDC-BERT} has the largest degradation when evaluated on the look-ahead scenario, with an average of $2.4\%$ accuracy drop across all setups, compared to $1.3\%$, $0.8\%$, and $1.0\%$ achieved by \emph{SDC}, \emph{ALL}, and \emph{ALL-SDC} respectively. 

Look-ahead is an important setup for an intent classifier because it supports intent prediction improvement while the dialogue is on-going.
More research is needed to determine the best way to integrate the unsupervised and semi-supervised tasks described here with the supervised data, to improve a discriminative model's performance.
We consider \emph{SDC-BERT}'s large degradation in this setup to be another incentive towards applying a unified test-to-test model, such as \emph{T5}, to the task.

\begin{table}[t]
\centering
\small
\begin{tabular}{@{}c@{ }| @{ }@{ }c@{ }@{ }|@{ }@{ }c@{ }@{ } || @{ }@{ }c@{ }@{ } | @{ }@{ }c@{ }@{ } || @{ }@{ }c@{ }@{ } | @{ }@{ }c@{ } }
 & \multicolumn{2}{@{}c}{\textit{MultiWOZ}} & \multicolumn{2}{@{}c}{\textit{SGD}} & \multicolumn{2}{c}{\textit{EDU}}   \\
 &\emph{3-u} & \emph{3-gen} & \emph{3-u} & \emph{3-gen} & \emph{3-u} & \emph{3-gen} \\
    \hline
     \emph{SDC-BERT} & 98.2 & 97.0 & 74.2 & 71.0 & +5.4 & +2.7 \\ 
      \hline
     \emph{SDC} & 96.6 & 95.5 & 73.6 & 72.6 & +6.3 & +4.6 \\
     \emph{ALL} & 98.2 & \textbf{97.9} & \textbf{76.6} & \textbf{75.3} & +6.3 & +5.5 \\
     \emph{ALL-SDC} & \textbf{98.7} & 96.6 & 74.6 & 72.6 & \textbf{+9.1} & \textbf{+10.0} \\
\end{tabular}
\caption{Results on four different intent classification data-sets.}
\label{tab:comparing bert performance to t5.}
\end{table}

\section{Hyper-Parameters, Configurations, and Experimental Details}
We use a single V100 GPU with 64 GB RAM in all experiments.
All T5-based models, namely \emph{SUC}, \emph{SDC}, \emph{ALL}, and \emph{ALL-SDC} share the same hyper-parameters with respect to their shared architecture and to the optimization process. They all use an initialized T5-Base \citep{DBLP:journals/corr/abs-1910-10683}. \emph{SDC-BERT} use an initialized BERT-Base encoder \citep{devlin2018bert} and share the same optimization process as the T5-based models. Configuration details and the hyper-parameters of the training process are provided in Table \ref{tab:models params}.

All hyper-parameters were tuned on the development set. We tuned the maximal sequence lengths \{128, 256, 512\} and the batch size \{32, 64\}. We then chose the best performing set of hyperparameter according to the higher accuracy score on the appropriate development set. For all models we set the amount of training epochs to be 50, with an early stopping criteria of 7. 

\begin{table}
\small
\centering
\begin{tabular}{l | l}
\toprule
\multicolumn{2}{c}{T5} \\
\hline
hidden size & 512 \\
number of attention heads & 8 \\
number of hidden layers & 6 \\
vocab size & 32128 \\
max sequence length & 256 \\
number of parameters & 220 M \\
\toprule
\multicolumn{2}{c}{Encoder - BERT} \\
\hline
hidden size & 768 \\
number of attention heads & 12 \\
number of hidden layers & 12 \\
vocab size & 30522 \\
hidden activation & \emph{'gelu'} \\
number of parameters & 108 M \\
\toprule
\multicolumn{2}{c}{Classifiers - Pooler} \\
\hline
input dim & 768 \\
first hidden dim & 768  \\
second hidden dim & 256 \\
output dim &  \#Intents \\
\toprule
\multicolumn{2}{c}{Optimization} \\
\hline
optimizer & \emph{AdamOptimizer} \\
beta1 & $0.9$ \\
beta2 & $0.997$ \\
epsilon & $1e-9$\\
batch size & 100 \\
\toprule
\end{tabular}
\caption{The hyper-parameters of the \emph{SUC}, \emph{SDC}, \emph{ALL}, \emph{ALL-SDC}, and \emph{SDC-BERT} models.}
\label{tab:models params}
\end{table}

\section{URLs of Code and Data}
We provide here the URLs for the datasets and code we have used:  

\begin{itemize}
    \item We use code and pre-trained weights of the pre-trained \emph{BERT-Base, Uncased} \citep{devlin2018bert} and the pre-trained \emph{T5} \citep{DBLP:journals/corr/abs-1910-10683} from huggingface:
    \url{https://github.com/huggingface/transformers}
    \item \emph{MultiWOZ} \citep{budzianowski2018multiwoz:2} conversations with intent labels are extracted from the official paper's repository: 
    \url{https://github.com/budzianowski/multiwoz}
    \item \emph{SGD} \citep{rastogi2019towards:9} conversations with intent labels are extracted from the official paper's repository: 
    \url{https://github.com/google-research-datasets/dstc8-schema-guided-dialogue}
\end{itemize}

\end{document}